\PassOptionsToPackage{table,dvipsnames}{xcolor}
\documentclass[11pt]{article}
\usepackage[numbers, square]{natbib}

\newcommand\Mycite[1]{%
  \citeauthor{#1}~(\citeyear{#1}) \cite{#1} }

\usepackage[T1]{fontenc}
\usepackage[utf8]{inputenc}
\usepackage{fullpage}
\usepackage{amsmath}
\usepackage{longtable}

\usepackage{wrapfig} 
\usepackage{hyperref}
\usepackage{amssymb}
\usepackage{multicol}
\usepackage[table,dvipsnames]{xcolor}
\usepackage{tabularx}
\usepackage{graphicx}
\usepackage{csquotes}
\usepackage{caption}
\usepackage{subcaption}

\definecolor{Or}{HTML}{ffe6cc}
\definecolor{Re}{HTML}{ecb3c6}
\definecolor{Vio}{HTML}{ccccff}
\definecolor{Gre}{HTML}{ccffcc}
\definecolor{Ye}{HTML}{fffbd2}

\usepackage{tikz}
\usetikzlibrary{shapes.geometric, arrows, decorations.pathreplacing,calligraphy}

\title{A Brain Age Residual Biomarker (BARB): Leveraging MRI-Based Models to Detect Latent Health Conditions in U.S. Veterans}

\usepackage{authblk}

\author[1]{Shahrzad Jamshidi}
\author[1]{Arthur Bousquet}
\author[1]{Sugata Banerji}
\author[2,3]{Mark F. Conneely}
\author[1]{Bita Aslrousta}
\affil[1]{Department of Mathematics and Computer Science, Lake Forest College, USA}
\affil[2]{Captain James A. Lovell Federal Health Care Center, USA}
\affil[3]{Rosalind Franklin University of Medicine and Science,USA}

\date{}

\begin{document}

%

\maketitle


\begin{abstract}
Age prediction using brain imaging, such as MRIs, has achieved promising results, with several studies identifying the model’s residual as a potential biomarker for chronic disease states. In this study, we developed a brain age predictive model using a dataset of 1,220 U.S. veterans (18–80 years) and convolutional neural networks (CNNs) trained on two-dimensional slices of axial T2 weighted fast spin-echo and T2 weighted fluid attenuated inversion recovery MRI images. The model, incorporating a degree-3 polynomial ensemble, achieved an $R^2$ of 0.816 on the testing set. Images were acquired at the level of the anterior commissure and the frontal horns of the lateral ventricles.

Residual analysis was performed to assess its potential as a biomarker for five ICD-coded conditions: hypertension (HTN), diabetes mellitus (DM), mild traumatic brain injury (mTBI), illicit substance abuse/dependence (SAD), and alcohol abuse/dependence (AAD). Residuals grouped by the number of ICD-coded conditions demonstrated different trends that were statistically significant ($p = 0.002$), suggesting a relationship between disease states and predicted brain age. This association was particularly pronounced in patients over 49 years, where negative residuals (indicating advanced brain aging) correlated with the presence of multiple ICD codes. These findings support the potential of residuals as biomarkers for detecting latent health conditions.
\end{abstract} 

\section{Introduction}

Recent research has demonstrated that brain imaging, such as MRIs, can predict chronological age with remarkable accuracy across various modalities. Studies employing {T1-weighted MRI images} \cite{Beheshti2019BiasAdjustment, Beck2021WhiteMatter, Dular2024T1wPreprocessing,  Korbmacher2024VersionShuffling, Li2024CrossStratified, ValdesHernandez2023Feasibility, Zhang2024BrainAge}, 
{T2-weighted MRI images} \cite{Hwang2021BrainAgeT2}, and {fMRI imaging} \cite{Beck2021WhiteMatter,Liem2017BrainAge}, among others \cite{An2024EEG, Chen2020TransferLearning, Cho2024CerebrovascularAge, Fang2024Comparative}, have consistently demonstrated the ability to model patient age effectively. Interestingly, many of these studies have observed that their model's residuals---the difference between actual and predicted age---may serve as proxies for latent disease states.

These residual-based markers appear under various names in the literature, such as \textit{BrainAGE} \cite{Franke2010BrainAgeEstimation}, \textit{brain-predicted age difference} \cite{Cole2018BrainAge}, and \textit{predicted age difference (PAD)} \cite{Chen2020TransferLearning, Jonsson2019BrainAge, Li2024CrossStratified}. For consistency, we will collectively refer to these biomarkers as \textbf{brain age residual biomarkers (BARBs)}. These biomarkers are increasingly recognized for their potential to provide insights into hard-to-diagnose health conditions that accelerate brain aging.

\subsection{Theoretical Framework for Residual Analysis}
From a statistical perspective, why might BARBs hold such significance? The squared residual of any predictive model is composed of three fundamental components: the bias of the model (the error introduced by approximating a complex real-world problem with a simplified model), the variance of the model (the variability of a model’s predictions with different training datasets), and the irreducible error \cite{ESL}. This relationship is captured by the well-known decomposition written in Equation \ref{eq:bvd}. 
\begin{equation} \label{eq:bvd}
\mathbb{E}[(y - \hat{f}(x))^2] = \underbrace{\big(f(x) - \mathbb{E}[\hat{f}(x)]\big)^2}_{\text{Bias}^2} + \underbrace{\mathbb{E}[\big(\hat{f}(x) - \mathbb{E}[\hat{f}(x)]\big)^2]}_{\text{Variance}} + \underbrace{\sigma^2}_{\text{Irreducible Error}}
\end{equation}
When considering the absolute value of the residual, a similar decomposition applies, though bias is not squared \cite{Gao2021MAEDecomposition}. This is often studied when considering a model’s ability to generalize (low bias) and its sensitivity to fluctuations in the training data (high variance), where these two concerns must be balanced--a concept known as the \textit{bias-variance trade-off}. Here, however, it is the
 {irreducible error} that warrants attention as external effects influences on the residual would reside. For a BARB to be maximally effective, the irreducible error must be an outsized portion of the residual and primarily influenced by the target health factor(s). 
 
\subsection{Clinical Applications and Evidence}

In the context of brain age prediction, the features are brain scans, often MRIs.  Sometimes, features include other health metrics like cognitive performance scores \cite{Liem2017BrainAge}. The labels are the chronological ages of  patients. The BARB is typically tested as a biomarker for some external factor(s), such as {hidden disease states}. Conditions like {Alzheimer's disease} \cite{Beheshti2019BiasAdjustment, Franke2010BrainAgeEstimation, Li2024CrossStratified} and {diabetes mellitus} \cite{Cole2018BrainAge, Zhang2024BrainAge} may cause the brain to \enquote{age} prematurely. Hence, if the predicted age is larger than the actual age of a patient, perhaps there is an underlying disease. As noted by \Mycite{Cole2018BrainAge}:
\begin{quote}
Brain structure is well-known to alter throughout life, and deviations from this typical brain aging trajectory, in terms of increased brain atrophy for a given age, may well reflect latent neuropathological influences.
\end{quote} 
BARBs are meant to capture this phenomenon.

\rowcolors{2}{Re!20}{Ye!20}

\scriptsize
\begin{longtable}{p{2.5cm}|p{2cm}|p{2cm}|p{2cm}|p{3cm}}
\rowcolor{Vio}
\textbf{Paper} & \textbf{Image Type} & \textbf{Participants} & \textbf{Source(s)} & \textbf{Model(s)} \\ \hline
\endfirsthead
\rowcolor{Vio}
\textbf{Paper} & \textbf{Image Type} & \textbf{Participants} & \textbf{Source(s)} & \textbf{Model(s)} \\ \hline
\endhead
\endfoot
\endlastfoot
\Mycite{Beck2021WhiteMatter} & T1-weighted MRI & 5,216 (45–86 years) & UK Biobank\cite{UKBiobank} & Linear Regression, Random Forest Regression, CNNs \\ 
\Mycite{Beheshti2019BiasAdjustment} & T1-weighted MRI & 1,698 (55–90 years) & ADNI\cite{ADNI} & Linear Regression with Bias Adjustment \\
\Mycite{Boyle2021BrainAge} & T1-weighted MRI & 1,359 training; external validation on datasets including Dokuz Eylül University, CR/RANN, TILDA & Multiple datasets (Dokuz Eylül University, CR/RANN, TILDA) & Elastic Net Regression \\
\Mycite{Chen2020TransferLearning} & Diffusion MRI & 1,380 (18–90 years) & CamCAN\cite{Shafto2014CamCAN, Taylor2017CamCAN}, NTUH, IXI\cite{IXIDataset} & Cascade Neural Network, Transfer Learning \\
\Mycite{Cho2024CerebrovascularAge} & MRA-based brain vessel images & 2,200 (ages unclear) & Own & Ensemble Model (RF, LSBoost, Linear Regression) \\
\Mycite{Cole2018BrainAge} & T1-weighted MRI & 2,670 (72–74 years) & LBC1936\cite{Deary2007LothianCohort1936, Deary2012LothianCohorts} & Gaussian Process Regression \\
\Mycite{Dartora2024BrainAge} & T1-weighted MRI & 15,289 (32–95.7 years) & ADNI\cite{ADNI}, AIBL\cite{Ellis2009AIBL}, GENIC, AddNeuroMed\cite{Lovestone2009AddNeuroMed}, J-ADNI\cite{JADNI}, UK Biobank\cite{UKBiobank} & CNN with ResNet architecture, using rigid MNI registration \\
\Mycite{Dular2024T1wPreprocessing} & T1-weighted MRI & 3,997 (45–81 years) & UK Biobank\cite{UKBiobank} & 4 CNNs (3D, 2D, Downsampled, Fully Convolutional) \\ 
\Mycite{Franke2010BrainAgeEstimation} & T1-weighted MRI & 650 training (19–86 years);  unknown & Own; ADNI\cite{ADNI} & Relevance Vector Regression \\
\Mycite{Hwang2021BrainAgeT2} & T2-weighted MRI & 1,230 participants (18–89 years) & Own & Deep Convolutional Neural Network (CNN) \\
\Mycite{Jonsson2019BrainAge} & T1-weighted MRI & 1,264 (18–75 years); 16,834 (46–80 years); 544 (20–86 years) & Own, UK Biobank \cite{UKBiobank}, IXI\cite{IXIDataset} & Residual CNN \\
\Mycite{Kianian2024GDSM} & T1-weighted MRI & 869 (20–86 years) & Own; IXI\cite{IXIDataset} & Greedy Dual-Stream Model (Local + Global CNNs) \\
\Mycite{Korbmacher2024VersionShuffling} & T1-weighted MRI & 4,395 (45–81 years) & UK Biobank\cite{UKBiobank} & Linear Regression, Version-Shuffling \\
\Mycite{Li2024CrossStratified} & T1-weighted MRI & 3,479 (18+ years) & SALD\cite{Wei2018SALD}, IXI\cite{IXIDataset}, NKI\cite{Nooner2012NKI, Tobe2022NKI}, DLBS\cite{Kennedy2015DLBS}, ICBM, CoRR\cite{Zuo2014CoRR}, CHCP\cite{Ge2023CHCP}, ADNI\cite{ADNI} & Cross-Stratified Ensemble (3D-DenseNet, ResNeXt, Inception-v4) \\
\Mycite{Liem2017BrainAge} & T1-weighted and functional connectivity MRI & 2,354 participants (19–82 years); additional 475 participants for validation & LIFE-Adult-Study\cite{loeffler2015life} & Multimodal CNN \\
\Mycite{ValdesHernandez2023Feasibility} & T1-weighted MRI & 840 (18–92 years) & UFHealth Dataset\cite{UFHealthDataset}; pretrained on UK Biobank\cite{UKBiobank} & DeepBrainNet (VGG16 CNN) \\
\Mycite{Vakli2024HeadMotion} & T1-weighted MRI & 34,099 (45–81 years) & UK Biobank\cite{UKBiobank}, MR-ART & 3D CNN, SFCN-reg \\
\Mycite{Zhang2024BrainAge} & T1-weighted, gray matter, and white matter & 1,716 (18–80 years) & Mixed Sources (unclear) & 3D Multi-Feature CNN (T1w, GM, WM) \\

\hiderowcolors\caption{A collection of studies using brain age residual biomarkers (BARBs). This table outlines the datasets, image types, and model architectures for each study. Significant representation is noted for convolutional neural networks (CNNs) and the use of T1-weighted MRI scans.}
\label{tab:datasets_models}
\end{longtable}
\normalsize

The reasoning behind using BARBs is fundamentally sound, with an established statistical framework underscoring their potential value. However, their effectiveness comes with notable caveats. The performance of a BARB depends heavily on the choice of model and the size and quality of the dataset, as these factors influence the balance between bias and variance \cite{ESL}. For instance, convolutional neural networks (CNNs), while powerful for image processing, are susceptible to overfitting (high variance), particularly when applied to small datasets. To mitigate this, studies like \Mycite{Cole2018BrainAge} and \Mycite{Dartora2024BrainAge} have implemented data augmentation strategies. However, these techniques introduce their own risks, such as reinforcing biases inherent in the dataset itself \cite{Goceri2023MedicalImageAugmentation}, thereby reducing generalizability. 
\begin{table}[h]
\centering
        \scriptsize
        \begin{tabular}{p{3cm}<{\raggedright}|p{2cm}|p{6cm}} 
        \showrowcolors
        \rowcolor{Vio}
        \textbf{Paper} & \textbf{Metrics} & \textbf{Values} \\ \hline
        \Mycite{Beck2021WhiteMatter} & MAE, $R^2$ & 6.99 years (MAE), 0.72 ($R^2$) \\ 
        \Mycite{Beheshti2019BiasAdjustment} & MAE, $R^2$ & 2.66 years (MAE), 0.81 ($R^2$) \\
        \Mycite{Boyle2021BrainAge} & MAE, $R^2$ & MAE: 5.7 years (training), 6.1 years (validation); $R^2$: 0.76 (training), 0.72 (validation) \\
        \Mycite{Chen2020TransferLearning} & MAE & 4.78 years (NTUH), 5.35 years (IXI-HH), 5.64 years (IXI-Guys) \\
        \Mycite{Cho2024CerebrovascularAge} & MAE & 6.818 years \\
        \Mycite{Cole2018BrainAge} & MAE, Hazard Ratio & MAE: 5.02 years (training), 7.08 years (testing); HR: 1.061 per year increase in Brain-PAD. \\
        \Mycite{Dartora2024BrainAge} & MAE, $R^2$ & 4.29 years (MAE), 0.94 ($R^2$) \\ 
        \Mycite{Dular2024T1wPreprocessing} & MAE, $R^2$ & 3.08 years (MAE), 0.97 ($R^2$) \\
        \Mycite{Franke2010BrainAgeEstimation} & MAE, $R$ & MAE: ~5 years; Correlation: \( R = 0.92 \). \\
        \Mycite{Hwang2021BrainAgeT2} & MAE, $R^2$ & 5.15 years (MAE), 0.80 ($R^2$) \\ 
        \Mycite{Jonsson2019BrainAge} & MAE, $R^2$ & 3.39 years (MAE), 0.87 ($R^2$) \\ 
        \Mycite{Kianian2024GDSM} & MAE & 3.25 years (Own), 4.18 years (IXI\cite{IXIDataset}) \\
        \Mycite{Korbmacher2024VersionShuffling} & MAE & 3.54 years \\        \Mycite{Li2024CrossStratified} & MAE, RMSE, $R^2$ & 2.94 years (MAE), 3.95 years (RMSE), 0.96 ($R^2$) \\
        \Mycite{Liem2017BrainAge} & MAE & 4.29 years (MAE) \\
        \Mycite{ValdesHernandez2023Feasibility} & MAE, $R^2$ & 8.05 years (MAE uncorrected), 0.33 ($R^2$ uncorrected), 0.76 ($R^2$ corrected) \\
        \Mycite{Zhang2024BrainAge} & MAE, PCC & 3.08 years (MAE), 0.97 (PCC) \\
        \end{tabular}
        \caption{Metrics and values from brain age prediction papers, alphabetized by the first author's last name. This table highlights the testing set metrics.}
        \label{tab:model_metrics}
\end{table}

Regression-based methods, some of which are less prone to overfitting, require preprocessing steps to make the image data compatible with the modeling framework. Dimensionality reduction techniques such as principal component analysis (PCA) have been used to address these issues by simplifying imaging data, enabling predictive models like support vector machines (SVMs) \cite{Cole2018BrainAge, Franke2010BrainAgeEstimation}. Yet, even preprocessing has its challenges. For example, \cite{Korbmacher2024VersionShuffling} demonstrated that version-dependent biases in preprocessing pipelines can significantly alter phenotype-brain age relationships. The challenges arising from the bias-variance trade-off are inescapable regardless of the models used given the complexity of images and the limited available data. 

The size of the dataset is not the only issue.  For example, imbalanced datasets have the potential to reduce a model's generalizability in brain age prediction. If there are too few subjects with a specific label, then that label will be poorly understood by the model. For this application specifically, it Is reasonable to presume that residuals may vary more with larger ages. Taking the age of the subjects within the dataset seems crucial for such an application and, indeed, several studies surveyed, such as \Mycite{Beheshti2019BiasAdjustment}, made adjustments for age-related biases. 

Table \ref{tab:datasets_models} summarizes several studies in terms of sample size, data sources, imaging modalities, and models used. Notably, about 70\% of the papers surveyed employ convolutional neural networks (CNNs), highlighting their dominance. This, however, does raise concerns regarding overfitting. The datasets informing these studies are predominantly sourced from the UK Biobank \cite{UKBiobank}, IXI \cite{IXIDataset}, and ADNI \cite{ADNI}, likely due to their high quality and availability. Still, ubiquitous use in this field may limit generalizability due to their demographic and geographic homogeneity.

Despite these challenges, the results summarized in Table \ref{tab:model_metrics} demonstrate that brain age prediction models achieve solid test metrics, even when mixed with ibid datasets, such as those used by \Mycite{Jonsson2019BrainAge} and others. Ensemble methods, which integrate architectures like DenseNet, ResNeXt, and Inception-v4, have shown particular promise in enhancing model performance by leveraging the complementary strengths of individual models \cite{Li2024CrossStratified}. By reducing total error, it is possible that ensembles may improve the informative power of a BARB. This suggests that building models to predict age using brain images is a sound learning task--something mirrored in our work here. Still, it is important to highlight that few papers contextualize test set metrics by comparing them to the training set. In \Mycite{Cole2018BrainAge}, this is provided and the better metrics with the training set suggest some overfitting.

The associated BARBs for these brain age models have been linked to a variety of disease states and health metrics. For example, they have been associated with diabetes, traumatic brain injury (TBI), schizophrenia, chronic pain, and neurodegeneration, including Alzheimer’s disease (AD) \cite{Jonsson2019BrainAge, Cole2018BrainAge, ValdesHernandez2023Feasibility}. Additionally, systemic health indicators such as weaker grip strength, slower walking speed, lower fluid intelligence, and higher allostatic load correlate with a BARB, alongside an increased risk of mortality \cite{Cole2018BrainAge}.

\begin{wrapfigure}{r}{0.65\textwidth} 
    \centering
    \resizebox{0.63\textwidth}{!}{

   \tikzstyle{block1} = [rectangle, draw, fill=purple!30,  
    text width=15em, text centered, rounded corners, node distance= 3cm, minimum height=3em] 

\tikzstyle{block} = [rectangle, draw, fill=blue!20,   
    text width=5em, text centered, rounded corners, node distance=2.25cm, minimum height=3em]  
 
 \tikzstyle{blockB} = [ellipse, draw, fill=yellow!20,   
    text width=2em, text centered, rounded corners, node distance=2.25cm, minimum height=3em]  
    
 \tikzstyle{blockC} = [ ellipse, draw, fill=green!20,   
    text width=2em, text centered, rounded corners, node distance=2.25cm, minimum height=3em]  

 \tikzstyle{pic} = [draw=blue!30, inner sep=-1mm, line width=2mm]  
 \tikzstyle{pic2} = [draw=purple!30, inner sep=-1mm, line width=2mm ]  
 
\tikzstyle{line} = [draw, -latex']  

\begin{tikzpicture}[node distance = 1.8cm, auto,ultra thick] 

\node[pic2] (fac) at (5.5,3.8)
    {\includegraphics[width=.30\textwidth]{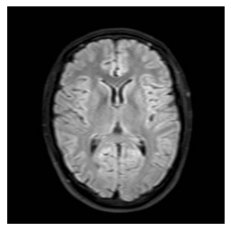}};
\node[pic] (flv) at (-0.5,-2.5)
    {\includegraphics[width=.30\textwidth]{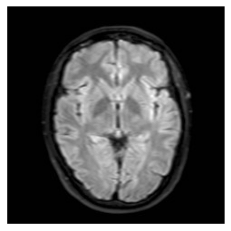}};
\node[pic2] (t2ac) at (9.5,-0.1)
    {\includegraphics[width=.30\textwidth]{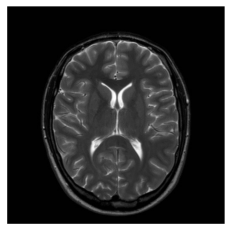}};
\node[pic] (t2lv) at (3.5,-6.5)
    {\includegraphics[width=.30\textwidth]{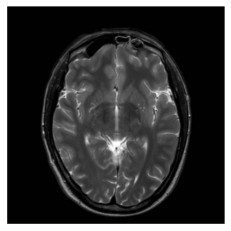}};
  
\node[inner sep=0] (lv) at (5.5,6.5) {Flair LV };
\node[inner sep=0] (lv) at (-0.5,0.2) {Flair AC };
\node[inner sep=0] (lv) at (9.5,2.6) {T2 LV };
\node[inner sep=0] (lv) at (3.5,-3.8) {T2 AC };

\node[inner sep=0] (S) at (4.5,-1.5)
    {\includegraphics[width=.20\textwidth]{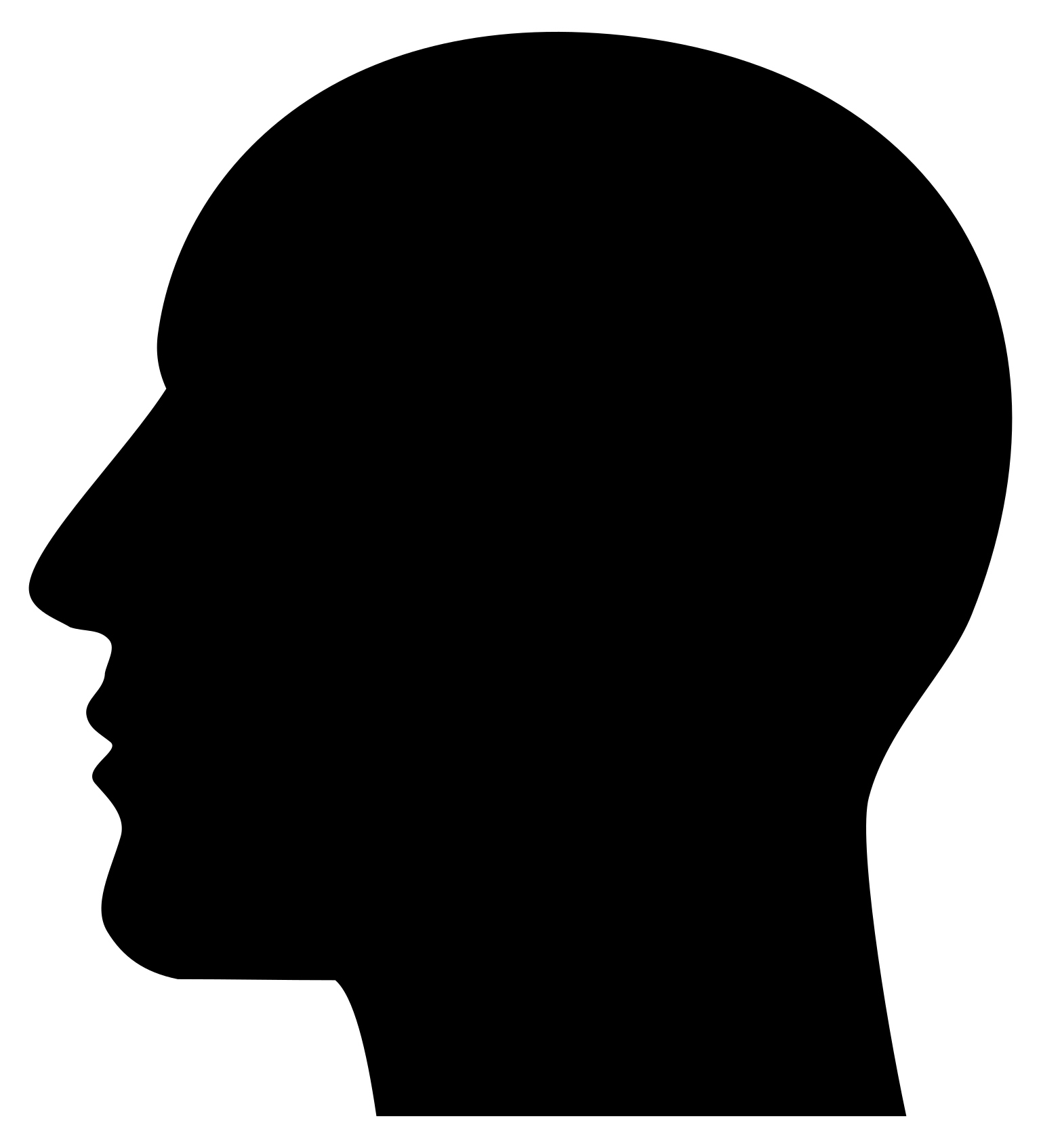}};
    
\path[color=purple!30, draw] (3,-0) -- (6.5,-1);
\path[color=blue!30, draw] (3,-0.3) -- (6.5,-1.3);

\end{tikzpicture}  
    
    }
    \caption{The model presented here is constructed from T2 weighted fast spin-echo (FSE) and T2 weighted fluid attenuated inversion recovery (FLAIR) images of two locations of the brain: the anterior commissure and the frontal horns of the lateral ventricles.}
    \label{fig:brains}
\end{wrapfigure}
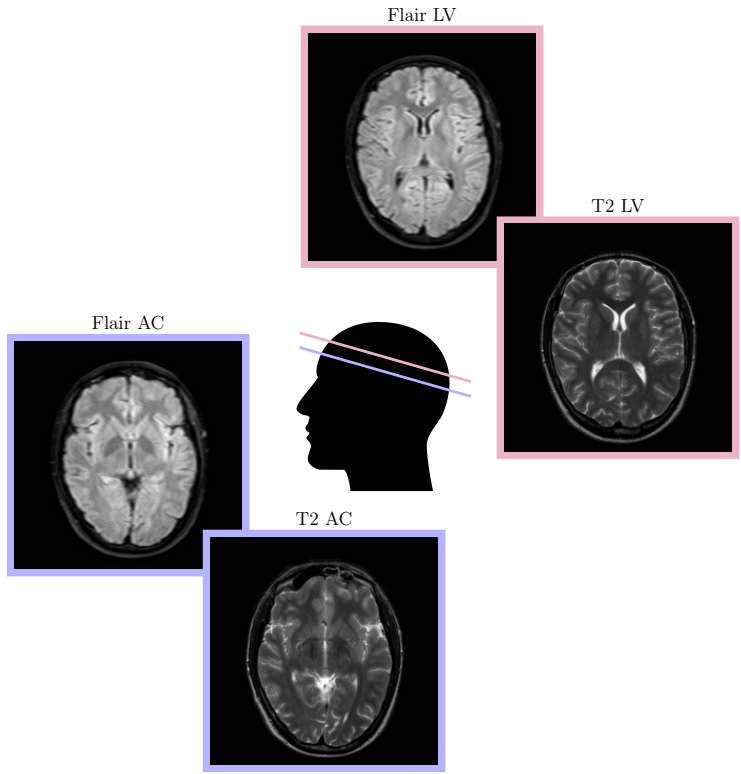

\pagebreak

\subsection{Research Focus}

This study contributes to the growing field of brain age residual biomarkers (BARBs) by utilizing a unique dataset of 1,220 U.S. veterans aged 20–80. Each participant contributed four MRI images, including axial T2 weighted fast spin-echo (FSE) and T2 weighted fluid attenuated inversion recovery (FLAIR) pulse sequences, captured at the level of the anterior commissure and at the level of the frontal horns of the lateral ventricles, as shown in Figure \ref{fig:brains}. To ensure reproducibility, images acquired through the frontal horns were obtained at the level of the greatest bifrontal diameter (distance between the lateral surface of the frontal horns).  T2 weighted images were used because they offer superior contrast resolution in separating CSF in the basal cisterns, sulci, and ventricles from brain tissue, making them ideal for capturing atrophy.  T2 weighted FLAIR images, on which CSF signal is suppressed, offered superior resolution of T2 hyperintense lesions seen in chronic small vessel ischemic disease of the cerebral white matter, the prevalence and extent of which correlate with patient age.  Much of the prior literature has relied on T1-weighted MRI scans, potentially due to their prevalence in large, widely used datasets.

To predict brain age, we construct four small convolutional neural network (CNN) models, each tailored to one of the four image types. The ensemble model integrates these predictions using a degree-3 polynomial regression without mixed terms. This mirrors the literature by utilizing both CNN and ensemble techniques, but departs in the actual design of the model. 

\begin{figure}
     \centering
\includegraphics[scale=0.44]{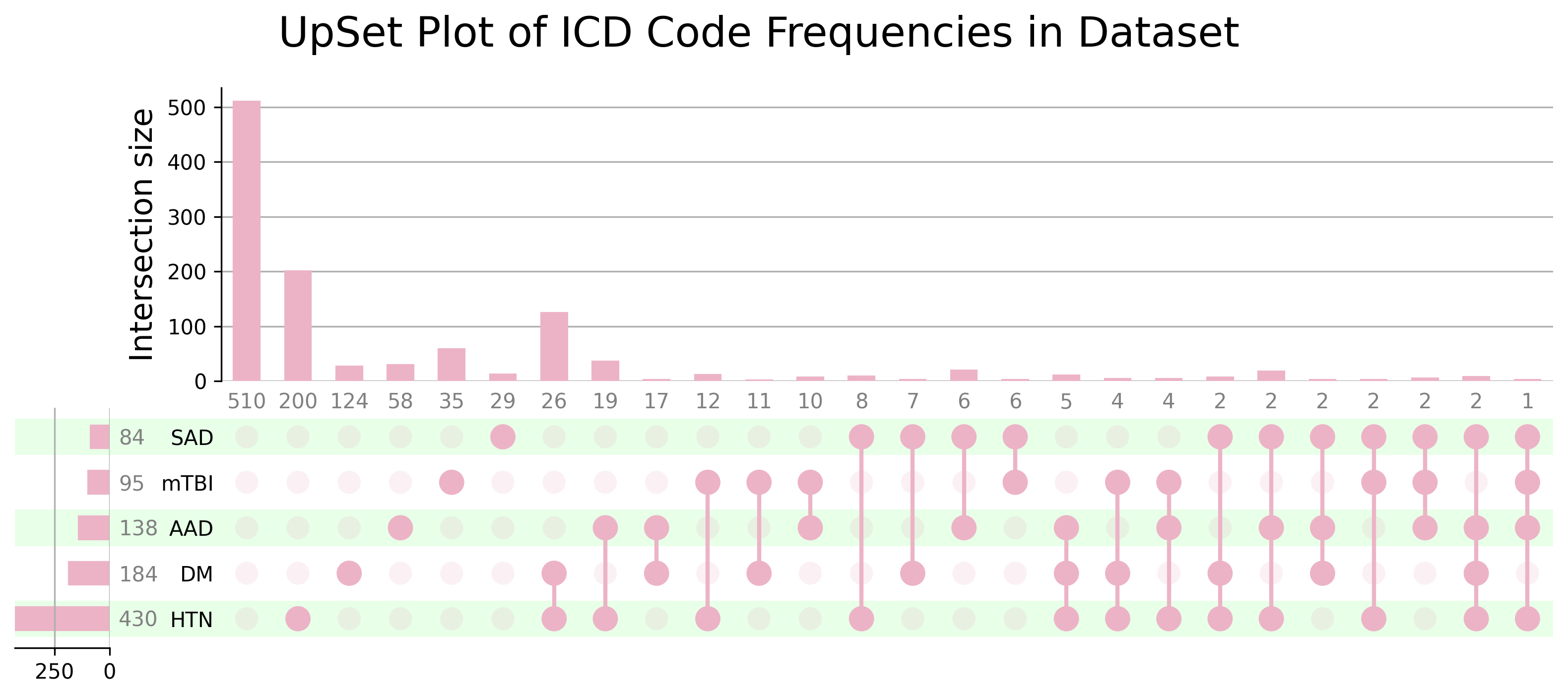}
        \caption{Distribution of the ICD codes, with a total of 1220 patients.}
        \label{fig:upset}
\end{figure}

The target BARB for this model focused on 5 challenging conditions: alcohol \\ abuse/dependence (AAD), substance abuse (SAD), hypertension (HTN), diabetes mellitus (DM), and mild traumatic brain injury (mTBI), see Figure \ref{fig:upset}. These conditions are often underreported, inconsistently managed, or clinically subtle. Hence these conditions would benefit from a BADB. Despite the promise of this approach, the results demonstrated only modest correlations between brain age residuals and these conditions. We surmise this is likely due to the statistical hurdles discussed earlier as well as challenges in labeling. For example, the binary labeling of these 5 ICD codes can lack the subtly of distinguishing between well-managed and poorly managed disease states. In the case of AAD and SAD, it seems possible and likely that several samples are mislabeled as lacking one or both disorders. 

The findings provide a meaningful insight: residuals where the predicted age exceeded the actual age beyond a specified threshold were more likely to correspond to the presence of ICD codes that face clear clinical challenges in terms of classification. This adds to the existing body of research that brain age models can capture clinically relevant signals. 

\section{Materials and Methods}

\subsection{Dataset}\label{sec:data}


The dataset used consisted of MRI scans from 1220 veterans, aged 20 to 80. For each age, there were 20 participants, with the exception of three ages: 62, 63, and 64, which had 23, 22, and 15 participants, respectively. The cohort was predominantly male, comprising 70 \% of the data. Aside from sex, no other demographic data was collected. 

Each participant had up to four 2D MRI images, selected by radiologists to ensure diagnostic relevance. These images were T2 weighted fast spin-echo (FSE) and T2 weighted fluid attenuated inversion recovery (FLAIR) sequences, taken from two key anatomical locations: the anterior commissure and the frontal horns of the lateral ventricles. These specific sequences were chosen for their ability to provide enhanced contrast between cerebrospinal fluid and gray matter to better detect aging. Moreover, these images were curated by radiologists rather than selected algorithmically. This increases the likelihood that the invariant captured by the model matches what has been observed by professionals and researchers in the clinical setting. A separate potential factor that can undermine the model is the quality of the pictures. In fact, \Mycite{Vakli2024HeadMotion} demonstrated movement could translate to a higher residual value. 

Additionally, the dataset includes five binary ICD codes for disease states, where 1 corresponds to a diagnosis within the patient's chart. The five conditions tracked are hypertension (HTN), diabetes mellitus (DM), mild traumatic brain injury (mTBI), illicit substance abuse/dependence (SAD), and alcohol abuse/dependence (AAD), which are observed as influencing brain structures \cite{Gasecki2013HypertensionBrain, LINGFORDHUGHES200542, OscarBerman2003AlcoholBrain, Moheet2015DiabetesBrain} except for mTBI, which has been show to be visible only after several injuries in animal models \cite{Desai2020MouseModelTBI}. Each of these codes presents unique challenges.  In the case of mTBI, SAD, and AAD are commonly hidden conditions that rely on patient disclosure. Both HTN and DM, extremely common in the population, exhibit wide variability in severity and management. So it is possible that a patient with an HTN diagnosis may be asymptomatic if the condition is well-managed, thereby having a positive residual (indicating a \enquote{younger} brain) while another patient may qualify as AAD but fail to disclose this information yet correspond to a negative residual (indicating an \enquote{older} brain). With this dataset, we have no way of determining this lurking factor. 

\begin{wrapfigure}{r}{0.6\textwidth} 
     \centering
\fbox{\includegraphics[scale=0.3]{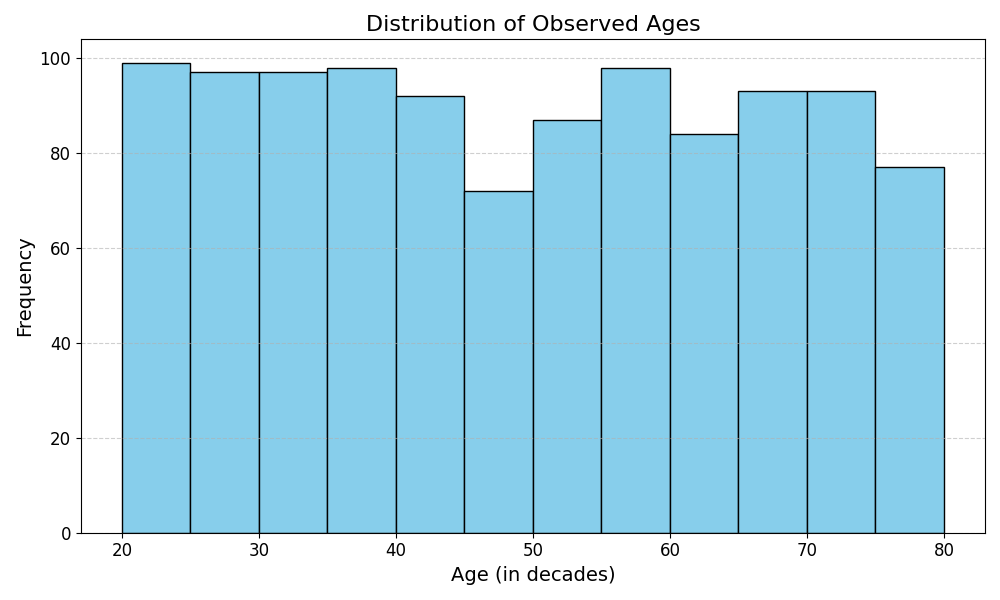}}
        \caption{Histogram illustrating the distribution of continuous ages in the reduced dataset of 1,104 subjects. Continuous age reflects fractional years, such as 26.25 for an individual aged 26 years and 90 days, providing a precise representation of age variability across the population.}
        \label{fig:histogram}
\end{wrapfigure}

Hence, predicting the ICD codes directly proves to be quite challenging. An effective brain age residual biomarker (BARB), however, should still correspond to a higher frequency of positive residuals in populations with these codes. This is under the assumption that on average, an \enquote{older} appearing brain is more likely to have at least one of these ICD codes. This aligns with the hypothesis that individuals with advanced brain aging (as indicated by positive residuals) are more likely to exhibit latent or overt disease states, even if we are unsure which disease state it might be.

Although data was collected on 1220 individuals, only 1104 had all four image typed in their file. This large subset was used in training, tuning and testing the ensemble model, while the remaining 116 were only used to tune the smaller CNNs. 

For the 1104 subjects, their ages were converted to continuous values, capturing fractional years, to improve the regression. Figure \ref{fig:histogram} illustrates the age distribution of the subset, which appears approximately uniform with slight overrepresentation in younger age groups (20–32) and underrepresentation in the 44–50 age range. The dataset captures a broad age range (20–80 years) with no extreme clustering, suggesting a relatively balanced dataset for analysis.

\begin{wraptable}{r}{0.7\textwidth}
    \scriptsize
    \centering
    \begin{tabular}{p{2.5cm}|p{6cm}}
        \rowcolor{Vio}
        \textbf{Model Type} & \textbf{Equation} \\ \hline
        Arithmetic Average &
        \(\tilde{y}=\frac{\tilde{y}_{1}+\tilde{y}_{2}+\tilde{y}_{3}+\tilde{y}_{4}}{4}\) \\
        Linear & 
        \(\tilde{y}=m_{1}\tilde{y}_{1}+m_{2}\tilde{y}_{2}+m_{3}\tilde{y}_{3}+m_{4}\tilde{y}_{4}+b\) \\
        Second Order &
        \(\begin{aligned}
        \tilde{y}=&m_{1}\tilde{y}_{1}+m_{2}\tilde{y}_{2}+m_{3}\tilde{y}_{3}+m_{4}\tilde{y}_{4}+ \\
                  & m_{5}y_{1}y_{2}+m_{6}y_{1}y_{3}+m_{7}y_{1}y_{4} +\\
                  & m_{8}y_{2}y_{3}+m_{9}y_{2}y_{4}+m_{10}y_{3}y_{4} +\\
                  & m_{11}y_{1}^2+m_{12}y_{2}^2+m_{13}y_{3}^2+m_{14}y_{4}^2+b
        \end{aligned}\) \\
        Third Order &
        \(\begin{aligned}
        \tilde{y}= & m_{1}\tilde{y}_{1}+m_{2}\tilde{y}_{2}+m_{3}\tilde{y}_{3}+m_{4}\tilde{y}_{4}+ \\
                  & m_{5}y_{1}y_{2}+m_{6}y_{1}y_{3}+m_{7}y_{1}y_{4}+\\
                  & m_{8}y_{2}y_{3}+m_{9}y_{2}y_{4}+m_{10}y_{3}y_{4}+\\
                  & m_{11}y_{1}^2+m_{12}y_{2}^2+m_{13}y_{3}^2+m_{14}y_{4}^2+\\
                  & m_{15}y_{1}^3+m_{16}y_{2}^3+m_{17}y_{3}^3+m_{18}y_{4}^3+b
        \end{aligned}\) \\
    \end{tabular}
    \caption{Four ensemble models were evaluated: arithmetic average, linear (which includes weighted-average), and second/third order models with higher-order terms. The final ensemble, selected via cross-validation, was a third-order model without interaction terms (Equation \ref{eq:ensemble}).}
    \label{tab:models}
\end{wraptable}

It is important to note that the dataset does not explicitly label brain images as \enquote{healthy}, which means the brain age model generates predictions based on what is typical within the dataset distribution rather than an established baseline of health.

To standardize the data, all scan images were preprocessed to ensure centering and resized to a resolution of $512 \times 512$. To enhance model training and mitigate overfitting, the training dataset was augmented by mirroring each image along the y-axis. While this augmentation increases the effective dataset size and helps prevent overfitting, it also introduces potential bias into the model, as mirrored images may reduce the independence of samples and artificially enhance symmetry in brain structures. Alternative data augmentation techniques, such as skull stripping and introducing slight deviations along the medial axis, were also explored. However, these methods did not yield significant improvements in model performance and were excluded from the final approach.

Additionally, we considered reframing the problem as a classification task by segmenting the age range into discrete bins. However, this approach was hindered by boundary noise, which led to poor performance metrics. For instance, the model struggled to reliably differentiate between individuals whose ages fell near the edges of adjacent categories, likely because being one year apart does not produce significant structural changes. This limitation further complicated the development of a robust brain age residual biomarker (BARB), as the inherent granularity of categorical predictions introduced unnecessary variability in the residuals. Hence, this was not included in the final approach as well.

\subsection{Model}

In this project, the predicted age was determined using an ensemble of four convolutional neural network (CNN) models, each trained on a specific image type (FSE or FLAIR sequence), as illustrated in Figure \ref{fig:brains}. The topology of the four CNN models is identical, with a batch size of 20 for training. Each model was designed to process grayscale images with input dimensions of $512 \times 512 \times 1$. The architecture consists of three convolutional layers with progressively increasing filter sizes (16, 32, and 64), each followed by batch normalization to stabilize training and improve model performance. Max pooling layers with a pooling size of $2 \times 2$ are applied after each convolutional block to reduce the spatial dimensions of the feature maps, ultimately producing a $62 \times 62 \times 64$ feature map. This feature map is flattened into a 1D array of size 246,016, which serves as input to the fully connected (dense) layers. The dense layers sequentially reduce dimensions, transitioning from 16 to 4 to 1 output, allowing the network to aggregate extracted features and produce a final prediction. A dropout layer with a threshold of 0.5 is included within the dense layers to mitigate overfitting by randomly deactivating neurons during training. In total, each CNN model contains 3,960,153 parameters, of which 3,959,897 are trainable on each image type.

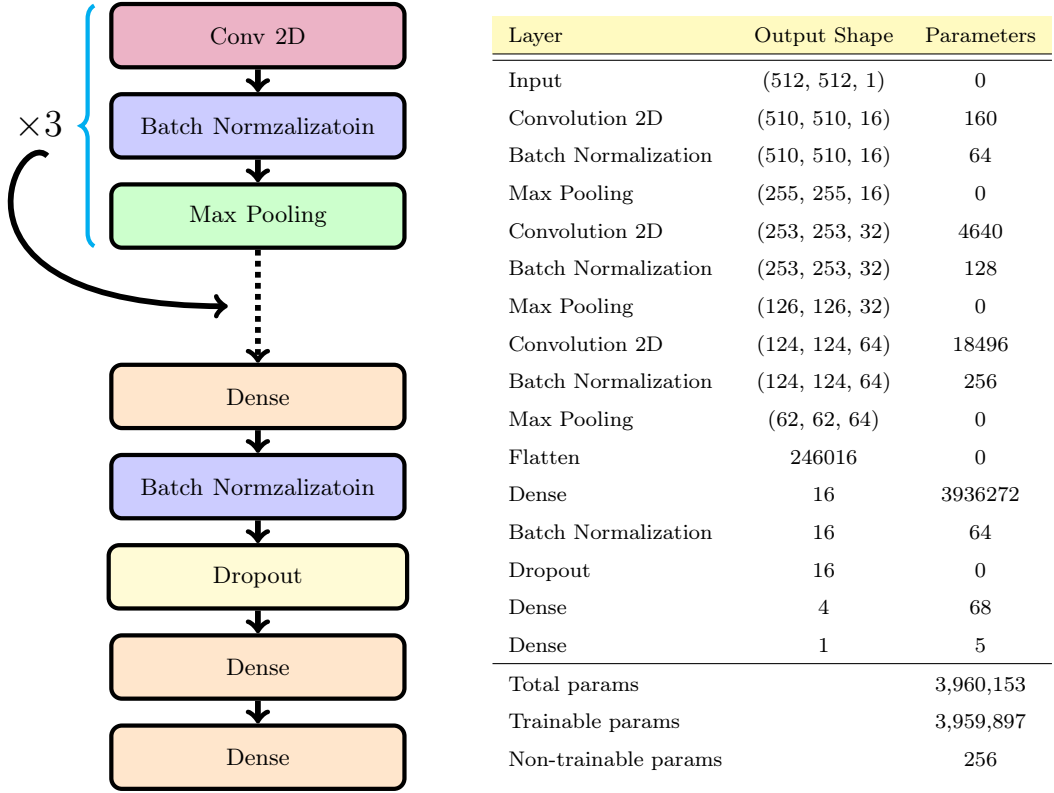
\begin{figure}
\begin{minipage}[l]{0.45\textwidth}
\begingroup
\tikzset{every picture/.style={scale=.7}}%

   \tikzstyle{convo} = [rectangle, draw, fill=purple!30,  
    text width=10em, text centered, rounded corners, node distance= 1cm, minimum height=2.2em, rotate=0, inner sep=1] 

\tikzstyle{batch} = [rectangle, draw, fill=blue!20,   
    text width=10em, text centered, rounded corners, node distance=1cm, minimum height=2.2em, rotate=0, inner sep=1]  

\tikzstyle{pool} = [rectangle, draw, fill=green!20,   
    text width=10em, text centered, rounded corners, node distance=1cm, minimum height=2.2em, rotate=0, inner sep=1]  

\tikzstyle{dense} = [rectangle, draw, fill=orange!20,   
    text width=10em, text centered, rounded corners, node distance=1cm, minimum height=2.2em, rotate=0, inner sep=1] 

\tikzstyle{dropout} = [rectangle, draw, fill=yellow!20,   
    text width=10em, text centered, rounded corners, node distance=1cm, minimum height=2.2em, rotate=0, inner sep=2]

\tikzstyle{line} = [draw, ->,  line width=0.2em]
\tikzstyle{dline} = [draw, dotted, ->,  line width=0.2em] 

\begin{tikzpicture}[node distance = 1.8cm, auto,ultra thick] 

\node [convo] (C1) at (0,0) {\footnotesize Conv 2D};
\node [batch](B1) at (0,-1.5) {\footnotesize Batch Normzalizatoin};
\node [pool](P1) at (0,-2*1.5) {\footnotesize Max Pooling};

\node [dense](D1) at (0,-4*1.5) {\footnotesize Dense};
\node [batch](B3) at (0,-5*1.5) {\footnotesize Batch Normzalizatoin};
\node [dropout](Dr1) at (0,-6*1.5) {\footnotesize Dropout};
\node [dense](D2) at (0,-7*1.5) {\footnotesize Dense};
\node [dense](D3) at (0,-8*1.5) {\footnotesize Dense};

\path[dline](P1)--(D1);
\path[line](C1)--(B1);
\path[line](B1)--(P1);
\path[line](D1)--(B3);
\path[line](B3)--(Dr1);
\path[line](Dr1)--(D2);
\path[line](D2)--(D3);

\draw [cyan,
    decorate, 
    decoration = {brace,mirror,
        raise=5pt,
        amplitude=5pt}] (-2.5,0.5) --  (-2.5,-2*1.5-0.5) node[pos=0pt,left=26pt,above=-55pt,black]{\Large $ \times 3 $} ;

\path[line] (-3.5,-2) .. controls +(-0.5,0.5)and +(-5,0).. (-0.5,-3*1.5);
      
\end{tikzpicture}  

\endgroup
\end{minipage} 
\begin{minipage}[l]{0.54\textwidth}
\scriptsize
\renewcommand{\arraystretch}{1.5}
\begin{tabular}{lcc}
\rowcolor{Vio} Layer           &     Output Shape       &       Parameters  \\
\hline
Input &  (512, 512, 1) & 0 \\
Convolution 2D       &  (510, 510, 16)    &  160       \\
Batch Normalization & (510, 510, 16)  &    64        \\
Max Pooling & (255, 255, 16)   &   0      \\
Convolution 2D &        (253, 253, 32)     & 4640      \\
Batch Normalization & (253, 253, 32)   &   128       \\
Max Pooling & (126, 126, 32)   &   0         \\
Convolution 2D &          (124, 124, 64)    &  18496     \\
Batch Normalization  & (124, 124, 64)   &   256\\
Max Pooling & (62, 62, 64)   &     0         \\
Flatten & 246016   &     0         \\
Dense & 16   &     3936272         \\
Batch Normalization & 16   &    64         \\
Dropout & 16   &     0         \\
Dense & 4   &     68         \\
Dense & 1  &     5         \\
\hline
Total params & & 3,960,153\\
Trainable params & &3,959,897\\
Non-trainable params & & 256\\
\end{tabular}
\end{minipage}
\caption{Architecture of the convolutional neural network (CNN) model used for each image-type. The layers are represented by horizontal bars, color-coded by type along with a table summarizing each layer. The batch size used as 20 and the pooling size was $2 \times 2$.} \label{fig:model}
\end{figure}

The CNN model designed for this study is a shallow, task-specific architecture optimized for predicting brain age from specific MRI data. Unlike deeper architectures such as ResNet and VGG16, which have been successfully applied to brain age prediction tasks \cite{Dartora2024BrainAge, ValdesHernandez2023Feasibility}, this model incorporates only three convolutional layers with progressively increasing filters (16, 32, 64), each followed by compact max pooling layers ($2 \times 2$). This streamlined design minimizes the risk of overfitting, a critical consideration given the small dataset size, but it inherently limits the model’s ability to learn complex hierarchical features. In contrast, ResNet leverages residual connections to enable efficient training of very deep networks, while VGG16 utilizes uniform $3 \times 3$ filters and large dense layers, resulting in high parameter counts and computational demands. When we implemented both ResNet and VGG16 for comparison, significant convergence issues were observed, particularly in training and testing metrics, underscoring the limitations of such deep architectures for small, specialized datasets.

\begin{figure}[h]
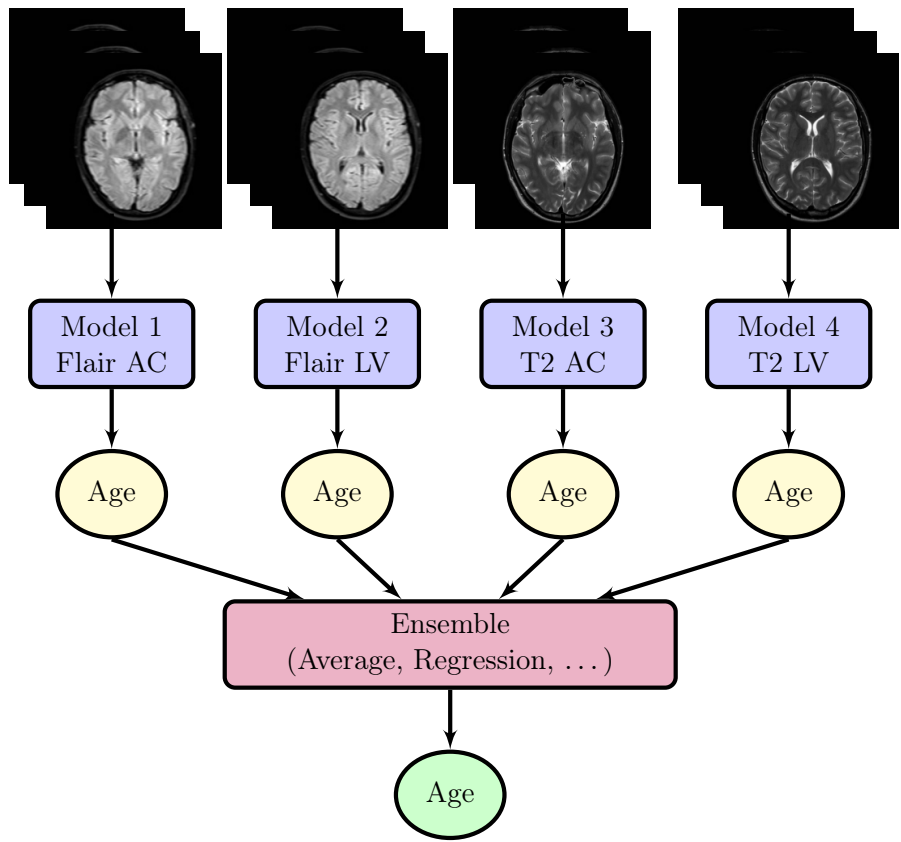

     \centering

   \tikzstyle{block1} = [rectangle, draw, fill=purple!30,  
    text width=15em, text centered, rounded corners, node distance= 3cm, minimum height=3em] 

\tikzstyle{block} = [rectangle, draw, fill=blue!20,   
    text width=5em, text centered, rounded corners, node distance=2.25cm, minimum height=3em]  
 
 \tikzstyle{blockB} = [ellipse, draw, fill=yellow!20,   
    text width=2em, text centered, rounded corners, node distance=2.25cm, minimum height=3em]  
    
 \tikzstyle{blockC} = [ ellipse, draw, fill=green!20,   
    text width=2em, text centered, rounded corners, node distance=2.25cm, minimum height=3em]  

\tikzstyle{line} = [draw, -latex']  

\begin{tikzpicture}[node distance = 1.8cm, auto,ultra thick] 

\node[inner sep=0pt] (fac) at (-0.2,3.3)
    {\includegraphics[width=.15\textwidth]{flair_ac_0.png}};
\node[inner sep=0pt] (fac2) at (0,3)
    {\includegraphics[width=.15\textwidth]{flair_ac_0.png}};
    \node[inner sep=0pt] (fac3) at (0.3,2.7)
    {\includegraphics[width=.15\textwidth]{flair_ac_0.png}};
    
\node[inner sep=0pt] (flv) at (2.7,3.3)
    {\includegraphics[width=.15\textwidth]{flair_lv_0.png}};
 \node[inner sep=0pt] (flv2) at (3,3)
    {\includegraphics[width=.15\textwidth]{flair_lv_0.png}};
\node[inner sep=0pt] (flv3) at (3.3,2.7)
    {\includegraphics[width=.15\textwidth]{flair_lv_0.png}}; 
    
\node[inner sep=0pt] (t2ac) at (5.7,3.3)
    {\includegraphics[width=.15\textwidth]{t2_ac_0.png}};
\node[inner sep=0pt] (t2ac2) at (6,3)
    {\includegraphics[width=.15\textwidth]{t2_ac_0.png}};
\node[inner sep=0pt] (t2ac3) at (6.0,2.7)
    {\includegraphics[width=.15\textwidth]{t2_ac_0.png}};

\node[inner sep=0pt] (t2lv) at (8.7,3.3)
    {\includegraphics[width=.15\textwidth]{t2_lv_0.png}};
\node[inner sep=0pt] (t2lv2) at (9,3)
    {\includegraphics[width=.15\textwidth]{t2_lv_0.png}};
\node[inner sep=0pt] (t2lv3) at (9.3,2.7)
    {\includegraphics[width=.15\textwidth]{t2_lv_0.png}};

\node [block] (A) at (0,0) {Model 1\\Flair AC};
\node [block](B) at (3,0) {Model 2\\Flair LV};
\node [block](C) at (6,0) {Model 3\\T2 AC};
\node [block](D)at (9,0){Model 4\\T2 LV};

\node [blockB] (ageA) at (0,-2) {Age};
\node [blockB](ageB) at (3,-2) {Age};
\node [blockB](ageC) at (6,-2) {Age};
\node [blockB](ageD)at (9,-2){Age};

    \node [block1](E) at (4.5, -4) {Ensemble\\(Average, Regression, $\dots$)};  
    
        \node [blockC](ageE) at (4.5, -6) {Age};  

 \path [line] (A) --(ageA); 
 \path [line] (B)--(ageB); 
 \path [line] (C)--(ageC); 
 \path [line] (D)--(ageD); 
 
 \path [line] (ageA.south) -- (E); 
 \path [line] (ageB.south) -- (E); 
 \path [line] (ageC.south) -- (E); 
 \path [line] (ageD.south) -- (E); 
 
 \path [line] (fac2) -- (A); 
\path [line] (flv2) -- (B); 
\path [line] (t2ac2) -- (C);
\path [line] (t2lv2) -- (D);

\path [line] (E) -- (ageE);
      
\end{tikzpicture}  

\caption{Four types of ensemble methods were evaluated for this study, as outlined in Table \ref{tab:models}. Ensemble approaches can reduce total prediction error, as described in Equation \ref{eq:bvd}, by leveraging the potential for error cancellation across the individual models.}
\label{fig:ensemble}
\end{figure}

After training each CNN model independently, we explored five types of ensembles: arithmetic mean, linear, second-order, and third-order models. Linear models allowed for weighted averages, while higher-order polynomial models incorporated potential interaction terms. However, coefficients for interaction terms in mixed models were statistically insignificant. Among the higher-degree models, the third-order polynomial demonstrated the best performance during cross-validation, as summarized in Table \ref{tab:cv}.

To formalize the ensemble framework, let $\hat{y}_{1}$, $\hat{y}_{2}$, $\hat{y}_{3}$, and $\hat{y}_{4}$ denote the predicted ages from the four CNN models trained on FLAIR AC, FLAIR LV, FSE AC, and FSE LV image types, respectively. The ensemble brain age prediction, $\hat{y}$, is represented by a third-order polynomial regression without interaction terms:

\begin{equation}
\hat{y} = 
    \underbrace{a_1\textcolor{red!80}{\hat{y}_{1}^3} + a_2\textcolor{blue!80}{\hat{y}_{2}^3} + a_3\textcolor{violet!80}{\hat{y}_{3}^3} + a_4\textcolor{cyan!80}{\hat{y}_{4}^3}}_{\text{Third-order terms}} 
    + \underbrace{b_1\textcolor{red!80}{\hat{y}_{1}^2} + b_2\textcolor{blue!80}{\hat{y}_{2}^2} + b_3\textcolor{violet!80}{\hat{y}_{3}^2} + b_4\textcolor{cyan!80}{\hat{y}_{4}^2}}_{\text{Second-order terms}} \\
    + \underbrace{c_1\textcolor{red!80}{\hat{y}_{1}} + c_2\textcolor{blue!80}{\hat{y}_{2}} + c_3\textcolor{violet!80}{\hat{y}_{3}} + c_4\textcolor{cyan!80}{\hat{y}_{4}}}_{\text{First-order terms}} 
    + \underbrace{d}_{\text{Intercept}}.
\end{equation}
\label{eq:ensemble}
\noindent where $a_i, b_i, c_i, d$ are weights trained on the data. The inclusion of first-order terms allows the ensemble to capture the baseline contributions of each CNN, while second-order terms reflect the inherent variability and nonlinear relationships within the models. The exclusion of interaction terms suggests that covariance between models does not significantly contribute to the predictive framework.

\rowcolors{2}{Ye!20}{Ye!20}

\begin{table}
\centering\scriptsize
\begin{tabular}{rccccc>{\columncolor{orange!10}}c>{\columncolor{orange!10}}c}
\rowcolor{Vio}
Model & split 1 & split 2   &  split 3   & split 4  & split 5   & Mean & Variance \\
\hline
\rowcolor{Re!80}
\multicolumn{8}{c}{$R^2$}\\
Linear & 0.78 & 0.79 & 0.82 & 0.80 & 0.82 & 0.798 & 0.0015 \\
Second Order & 0.78 & 0.81 & 0.81 & 0.82 & 0.81 & 0.806 & 0.0003 \\
\rowcolor{Gre!90}
Third Order & 0.80 & 0.79 & 0.82 & 0.82 & 0.81 & 0.808 & 0.0003 \\
Fourth Order & 0.79 & 0.79 & 0.82 & 0.81 & 0.81 & 0.804 & 0.0003 \\
Average & 0.75 & 0.67 & 0.74 & 0.76 & 0.65 & 0.714 & 0.0029 \\
\rowcolor{Re!80}
\multicolumn{8}{c}{MAE}\\
Linear & 5.82 & 5.77 & 5.07 & 5.43 & 5.42 & 5.502 & 0.094 \\
Second Order & 5.80 & 5.76 & 5.14 & 5.20 & 5.55 & 5.490 & 0.078 \\
\rowcolor{Gre!90}
Third Order & 5.49 & 5.64 & 5.12 & 5.18 & 5.56 & 5.398 & 0.046 \\
Fourth Order & 5.58 & 5.60 & 5.15 & 5.17 & 5.54 & 5.408 & 0.031 \\
Average & 5.81 & 6.56 & 5.71 & 5.68 & 6.46 & 5.844 & 0.099 \\
\rowcolor{Re!80}
\multicolumn{8}{c}{MSE}\\
Linear & 55.94 & 52.31 & 41.66 & 52.68 & 47.54 & 50.026 & 30.156 \\
Second Order & 56.23 & 52.76 & 41.99 & 49.70 & 49.07 & 49.950 & 27.154 \\
\rowcolor{Gre!90}
Third Order & 51.31 & 52.41 & 41.06 & 47.47 & 49.49 & 48.748 & 19.806 \\
Fourth Order & 52.28 & 51.97 & 41.87 & 47.46 & 49.19 & 48.554 & 19.136 \\
Average & 55.18 & 63.49 & 49.73 & 52.24 & 60.58 & 56.244 & 31.537 \\
\end{tabular}
\caption{Metrics from cross-validation of various ensemble models used in the study, with additional variance calculations. This table compares the performance of arithmetic mean, linear, second-order, and third-order models without interaction terms. Interaction terms were excluded due to coefficients being statistically indistinguishable from zero, highlighting their negligible contribution. The third-order model was selected, although it is important to note that the metrics are very close.}
\label{tab:cv}\normalsize
\end{table}

The presence of meaningful third-order terms suggests that skewness in age predictions plays a significant role, a phenomenon observed in other studies that made explicit adjustments for age to account for increased variability. The ensemble method applied here inherently addresses such biases and higher-order effects by leveraging a polynomial framework that autonomously captures nonlinear patterns and asymmetries in the predictions without manual intervention.

Given that a third order model was successful, the inclusion of sex as a variable was explored to determine whether it could also improve model performance. However, the best-performing models did not incorporate sex, suggesting that this factor did not contribute significantly.

As detailed in Table \ref{tab:cv}, four ensemble types were evaluated, assuming no interaction terms and using the CNN models outlined in Figure \ref{fig:model}. The third-order polynomial was selected as the final model due to its favorable metrics and lower variability across cross-validation splits. While the fourth-order model also demonstrated strong performance, it did not provide a substantial improvement over the third-order model and was therefore excluded to favor a less complex design.

\section{Results}
\rowcolors{2}{Re!20}{Ye!20}
\begin{wraptable}{r}{0.4\textwidth}
\centering
\scriptsize 
\begin{tabular}{l|c}
\rowcolor{Vio}
\textbf{Test Metric}      & \textbf{Value}       \\ \hline
$R^2$      & 0.816                \\ 
Mean Absolute Error (MAE) & 5.450           \\ 
Mean Squared Error (MSE) & 48.348 \\ 
\end{tabular}
\caption{Model test set metrics.}
\label{tab:performance}
\end{wraptable}

Out of the 1104 samples with all four images, we used 883 for training and 221 testing. Since we augment the training set by flipping every image, our training set consists of 1766 images for each CNN model. No validation set was used; instead 5-fold cross validation was used for model selection as is implied by Table \ref{tab:cv}.

\subsection{Model Performance}

The ensemble model achieved an $R^2$ of 0.816 on the test set, a mean absolute error (MAE) of 5.450, and a mean squared error (MSE) of 48.348, summarized in Table \ref{tab:performance}. This performance compares favorably to models in the literature, exceeding the $R^2$ reported by studies such as \Mycite{Boyle2021BrainAge} and \Mycite{Jonsson2019BrainAge}. Although the MAE is higher than many of the models reviewed, such as \Mycite{Dular2024T1wPreprocessing}, it is still within the range of metrics observed in Table \ref{tab:model_metrics} of roughly 3 to 8 years.

It is important to note that the cross-validation metrics are on-par with those observed in the test set. The training MAE was similar, but the training $R^2$ metric was more exuberant (0.97). This means we do have mild evidence of overfitting to the data--a central concern when using a CNN-based model on a small dataset. The studies reviewed for this project omitted this important comparison with the exception of \Mycite{Cole2018BrainAge}, which did demonstrate evidence of overfitting as well. As a result, it is difficult to assess how this compares to the literature. 

Overall, it appears we have a reasonable model for predicting brain age with only mild evidence of variability (see Equation \ref{eq:bvd}).

\begin{wrapfigure}{l}{0.7\textwidth}
\centering
\includegraphics[scale=0.3]{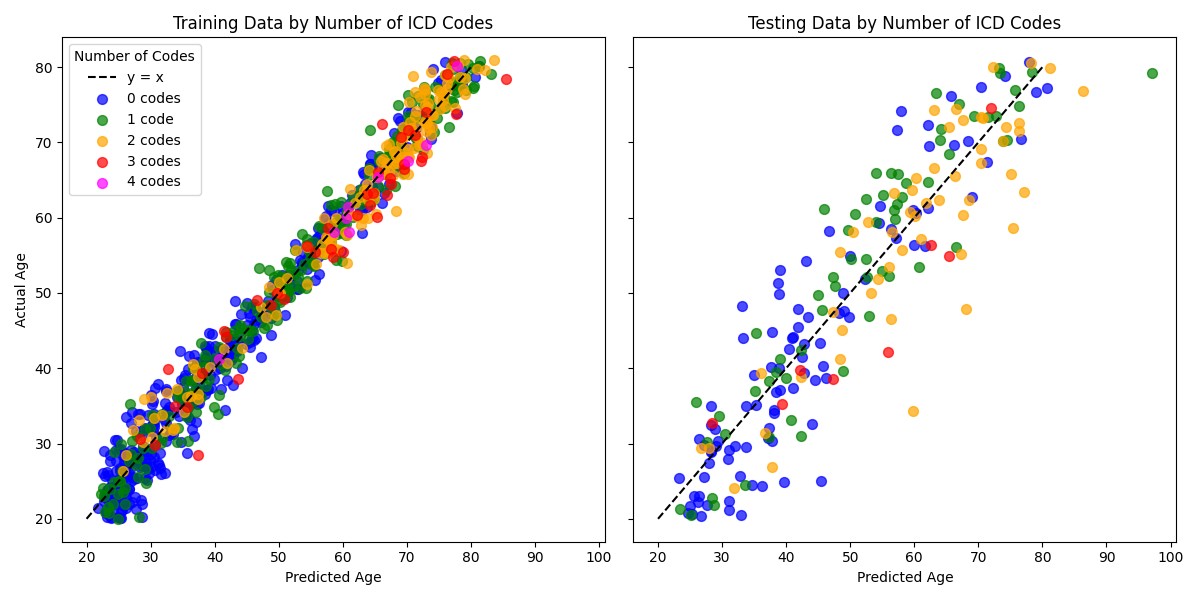}
\caption{Predicted age vs. real age for train and test sets, color-coded by number of ICD codes (HTN, DM, mTBI, SAD, AAD). Deviations from the diagonal (y = x) reflect residuals.}
\label{fig:residuals_by_ICD}
\end{wrapfigure}

\subsection{Residual Analysis}

An analysis of the residuals is the natural next step to assess the potential of a BARB using this model, which not only relies on different architecture but also different data. Figure \ref{fig:residuals_by_ICD} depicts the plot of real and predicted ages. The plot demonstrates strong alignment along the diagonal, indicating a high degree of accuracy in the model’s predictions. The variability is minimal and relatively uniform, suggesting that the model effectively captures underlying patterns in the data. Very mild evidence of overfitting exists as the test set exhibits  greater variability in residual values compared to the training set. Overall, these plots lend credibility to the brain age model constructed here.

Recall, however, the dataset’s lack of distinction for \enquote{healthy} brains: this will introduce a baseline error. Even if the model is well constructed and the residuals lack an outsized influence of model bias and/or variance, the baseline may not correspond to right target state, depending on the data, making it more challenging to detect disease states. It is therefore not surprising that no significant associations were found for DM, mTBI, or AAD in mean residuals. An ANOVA test did reveal significant differences in mean residuals for HTN and SAD ($p = 0.000$). Notably, HTN displayed an inverse correlation with the brain age residual, indicating younger predicted brain ages relative to chronological age ($p = 0.000$). This inverse relationship may reflect variability in the neurological effects of HTN based on factors such as disease management and severity, which the dataset does not differentiate. When assessing the positive correlation between SAD and brain age residuals, it is possible that patients with more severe cases are more likely to volunteer this information but this is only speculation.

A key limitation of the dataset is the binary nature of the ICD codes, which indicates only the presence or absence of a condition without additional details. For example, DM is not supplemented with A1C levels, and HTN lacks systolic or diastolic measurements. We have no way of assessing if individuals with well-managed conditions may exhibit different neurological effects compared to those with poorly managed conditions, because both are assigned the same ICD code. Further complicating the interpretation of results is the reliance on self-reported information for mTBI, SAD, and AAD. This reliance introduces the possibility of underreporting, as individuals meeting the criteria for these conditions may not disclose them. Recording inconsistencies may also exist, with some patients meeting ICD code criteria inaccurately labeled in the dataset. These factors, especially combined with the baseline issue, collectively complicate the understanding of this model residual's correlation with specific disease states.

\begin{wrapfigure}{l}{0.7\textwidth}
\centering
\includegraphics[scale=0.25]{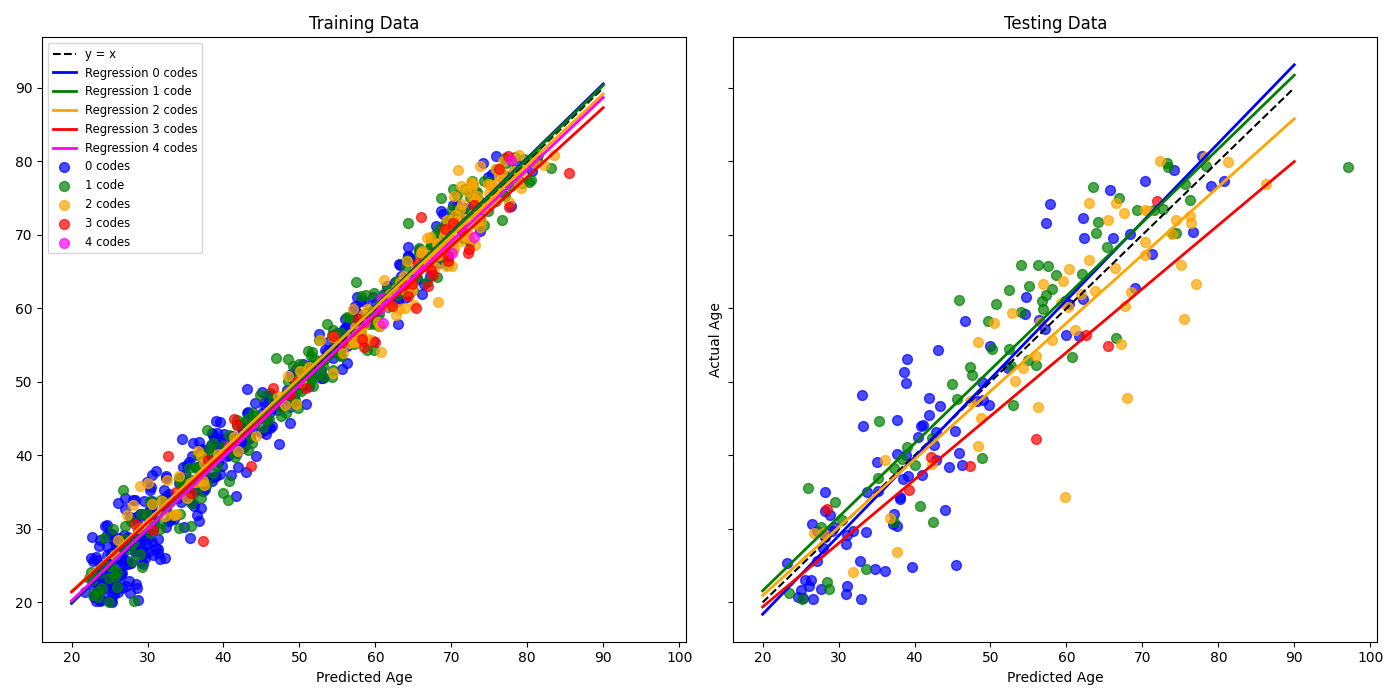}
\caption{Predicted age vs. real age for train and test sets, color-coded by number of ICD codes with trend lines added. Trendlines of the groups within the training set demonstrate a statistically significant difference (F-statistic 4.265, p-value 0.002). Out of curiosity, we verified the testing set had similar results (F-statistics 16.4, p-value 5.77e-13).}
\label{fig:trend_lines}
\end{wrapfigure}

Although the dataset lacks granularity and may be inconsistent with labels of these five disease states, which confounds direct prediction of the specific ICD codes and produces conflicting results with ANOVA tests, we can still assess if there is a correspondence with the model residual and the disease states. Since labeling inconsistencies and misclassification are assumed to occur randomly across the dataset, any systematic relationship between negative residuals—-where the predicted brain age exceeds the chronological age—-and the presence of ICD codes would provide evidence of a meaningful correlation. Figure \ref{fig:trend_lines} captures exactly this. When organizing subjects in the training set by how many ICD codes are associated to them, we see each group's trend line to be distinct ($p=0.002$). This means that given an actual age and the number of ICD codes, the predicted age will change. In particular, those with two or more trend toward an older brain age.

The direct analysis of the training residuals grouped by number of ICD codes provided in Table \ref{tab:training_residuals} reveals distinct patterns across groups. Mean residuals show a slight shift from positive values (younger predicted brain ages) for individuals with no ICD codes to negative values (older predicted brain ages) for those with higher ICD counts. Median residuals follow a similar trend, reinforcing this observation. Skewness values are near zero for most groups, suggesting generally symmetric residual distributions, with slight positive skewness observed for individuals with zero or one ICD code. In other words, these two groups exhibit a tail on the positive side (associated with younger predicted brain ages). Groups with two ICD codes exhibit moderate positive skewness, reflecting a greater likelihood of outliers of those with younger predicted brain ages. Kurtosis values reveal that individuals with one or two ICD codes have heavier-tailed distributions, while those with zero, three, or four ICD codes show flatter, more normal-like distributions. These findings suggest a nuanced relationship between predicted brain age residuals and the number of ICD codes, where residual variability may increase with the presences of 1 or 2 ICD codes. Still, the trends toward an older predicted brain age is evident, mirroring the trend lines seen in Figure \ref{fig:trend_lines}.
\begin{wraptable}{r}{0.63\textwidth}
    \centering
    \scriptsize
    \begin{tabular}{|l|c|c|c|c|c|}
        \hline
        \rowcolor{Vio} \textbf{Metrics} & \textbf{0 codes} & \textbf{1 code} & \textbf{2 codes} & \textbf{3 codes} & \textbf{4 codes} \\ 
        \hline
        Mean Residual  & 0.07 & 0.21 & 0.12 & -0.82 & -0.72 \\ 
        \hline
        Median Residual & -0.08 & 0.10 & -0.06 & -0.71 & -0.44 \\ 
        \hline
        Skewness        & 0.0397 & 0.0764 & 0.2207 & 0.0466 & -0.0014 \\ 
        \hline
        Kurtosis        & -0.0271 & 0.5321 & 0.4954 & 0.0282 & -1.0238 \\ 
        \hline
    \end{tabular}
    \caption{Statistics for residuals in the training set by the number of ICD codes. Metrics include mean, median, skewness, and kurtosis for residuals corresponding to 0, 1, 2, 3, and 4 ICD codes.}
    \label{tab:training_residuals}
\end{wraptable}

\begin{wrapfigure}{l}{0.7\textwidth}
\centering
\includegraphics[scale=0.25]{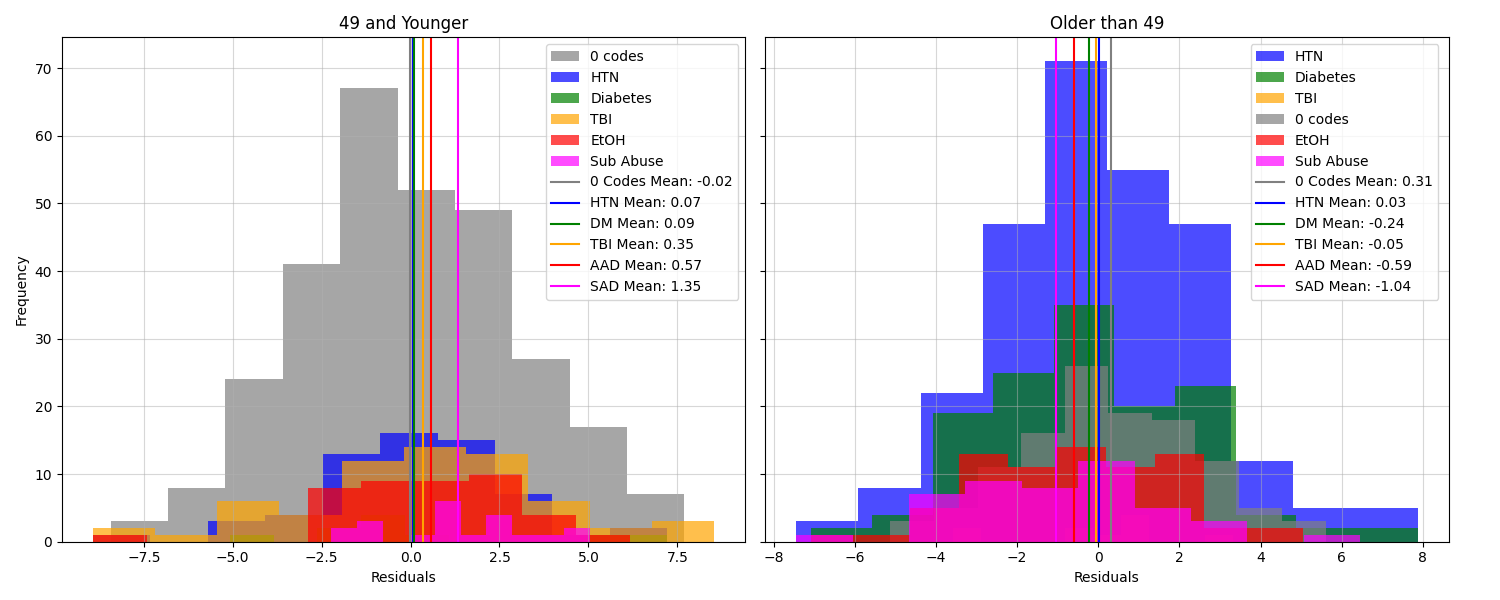}
\caption{Histograms of residuals broken down by ICD codes (see Figure \ref{fig:upset} for overlaps). Negative residuals correspond to a predicted brain age that is older than the actual age.}
\label{fig:50Hist}
\end{wrapfigure}

The analysis indicates that the number of ICD codes influences the residuals, yet attempts to model the reverse direction—predicting the number of ICD codes based on patient age and residuals—were unsuccessful. The signal-to-noise ratio appears insufficient, suggesting that numerous unobserved factors impact the ICD codes, making prediction unreliable with the available features. Among the models tested, decision trees consistently split the dataset at an actual age of approximately 48 or 49, depending on the data representation. This split is supported by Figure \ref{fig:residuals_by_ICD}, which reveals a sharp decline in the proportion of subjects with zero ICD codes after this age threshold.

Statistical analysis corroborates this observation. A one-way ANOVA test of residuals grouped by the number of ICD codes did not yield statistically significant differences across the groups ($p = 0.1880$). However, restricting the analysis to subjects older than 49 revealed significant differences ($p = 0.0069$). Further exploration of this age-based split, as shown in Figure \ref{fig:50Hist}, highlights that residual means are positive (indicating a younger predicted brain age) for subjects aged 49 or younger, regardless of the specific ICD code. Notably, SAD exhibits the highest mean residual within this subgroup, though this result may be a statistical artifact due to the small sample size. Conversely, for subjects older than 49, residual means generally become negative, except for those with no associated ICD codes or those with HTN.

These findings suggest that while the five ICD codes exert a detectable influence on the residuals, their effects are secondary to the dominant influence of the patient’s actual age. Despite these limitations, the ability to detect these subtle associations in a dataset with inherent challenges and deviations from standard literature strengthens the case for BARBs as a promising area of future research.

\section{Discussion}

This study explores alternative methodologies and datasets, departing from conventional practices in the literature in several key ways. Unlike the dominant reliance on T1-weighted MRI scans and large-scale datasets, this work utilized two specific image modalities—FLAIR and FSE MRI slices—and a dataset of 1220 U.S. veterans from a regional demographic. The ensemble model developed for this study consisted of lightweight convolutional neural networks (CNNs) and a third-order polynomial regression ensemble, which effectively captures mean, variance, and skewness in the CNN models as well as reduce the total error via cancellation. Additional analysis of the coefficients derived from the ensemble model, informed by principles such as those outlined in \cite{Joshi2010ThirdOrderBehavior}, could shed light on the specific contributions of higher-order terms and guide model refinements in future work. Despite these differences, the model achieved results comparable to existing benchmarks, with a competitive $R^2$ score of 0.816 and a high but reasonable mean absolute error (MAE) of 5.450. These findings underscore the efficacy of using alternative image modalities and ensemble strategies to capture meaningful patterns in smaller, specialized datasets. This also provides evidence for broader application that what has been previously seen. 

Several limitations of this study warrant discussion. The dataset, while unique, did not differentiate between healthy and unhealthy brain images, potentially introducing biases in the baseline of the brain age model and mitigating the model's predictive ability for disease states. Moreover, the binary representation of ICD code obscures important details of underlying health conditions, like  severity and management, that may translate into real neurological effects. The relatively small sample size ($n=1104$) coupled with the high complexity of processing image data, sets the learning problem up for high variability in the model error. Providing analysis and metrics indicating the extent of overfitting seems crucial for this domain.

Future research should aim to address these limitations by devising strategies for defining a baseline for such models. Additional imaging modalities and advanced augmentation techniques should be explored to improve the generalizability of the model. Refining the representation of ICD codes to include severity metrics, such as A1C levels for DM or blood pressure measurements for HTN, could enhance the interpretability of findings. Finally, incorporating other potential biomarkers as features, something seen in the literature, could strengthen findings significantly.

The residual analysis did not find compelling associations between the model errors and any one of the five ICD codes considered: substance abuse/dependence (SAD), alcohol abuse/dependence (AAD),  mild traumatic brain injury (mTBI), diabetes mellitus (DM) and hypertension (HTN). This may have been due to labeling issues. In aggregate, the labels had a compelling association with negative residuals (an older predicted brain age), which became definitive when we considered If the patient was older than 49. These findings provide evidence that the residuals, or the difference between predicted and actual brain age, may serve as potential brain age residual biomarkers (BARBs) for identifying subtle or latent health conditions.  

In this work , we've discussed why BARBs are theoretically well-founded in statistics and that that their potential is evident even in challenging circumstances. The ability of BARBs to capture nuanced relationships between brain aging and systemic health conditions offers promise for non-invasive screening and monitoring for hard-to-diagnose disease states. This approach could be particularly valuable in identifying early indicators of diseases with subtle or hidden symptoms, providing actionable insights for clinicians. While the model presented here demonstrates potential, there appears to be many different effective models evident in the literature. The development for a clinically applicable BARB rests on on the need for a large, more diverse dataset with an array of image types and several accompanying health metrics, a massive undertaking.

%
%
%

\bibliographystyle{plainnat}

\end{document}